\documentclass{article}

\PassOptionsToPackage{numbers,compress}{natbib}
\usepackage[preprint]{neurips_2025}

\usepackage[utf8]{inputenc} % allow utf-8 input
\usepackage{graphicx}
\usepackage{amsmath, amssymb}
\usepackage[ruled,vlined]{algorithm2e}
\usepackage[T1]{fontenc}    % use 8-bit T1 fonts
\usepackage{hyperref} 
\usepackage{multirow}
\usepackage{url}            % simple URL typesetting
\usepackage{booktabs}       % professional-quality tables
\usepackage{amsfonts}       % blackboard math symbols
\usepackage{nicefrac}       % compact symbols for 1/2, etc.
\usepackage{microtype}      % microtypography
\usepackage{xcolor}         % colors
\usepackage{algorithmic}
\usepackage{amsmath}
\usepackage{xspace}
\usepackage{subcaption}
\usepackage{tabularx}   % <-- defines tabularx

\newcommand{\method}{\textsc{RAxSS}\xspace}

\title{RAxSS: Retrieval-Augmented Sparse Sampling for Explainable Variable-Length Medical Time Series Classification}
\workshoptitle{Learning from Time Series for Health
}

\author{%
  Aydin Javadov \\
  ETH Zurich \\
  \texttt{ajavadov@ethz.ch} \\
  \\
  \And
  Samir Garibov \\
  University of Freiburg \\
  \texttt{garibovs@cs.uni-freiburg.de} \\
  \And
   Tobias Hoesli \\
   ETH Zurich  \\
    \texttt{thoesli@ethz.ch} \\
  \And
  Qiyang Sun \\
  Imperial College London \\
  \texttt{q.sun23@imperial.ac.uk} \\
  \And
  Florian von Wangenheim \\
  ETH Zurich \\
  \texttt{fwangenheim@ethz.ch} \\
  \And
  Joseph Ollier 
  \\
  ETH Zurich \\
  \texttt{jollier@ethz.ch} \\
   \And
  Björn Schuller
  \\
  Imperial College London \& \\ Technical University of Munich \\
  \texttt{bjoern.schuller@imperial.ac.uk}
   \\
}

\begin{document}

\maketitle

\begin{abstract}

Medical time series analysis is challenging due to data sparsity, noise, and highly variable recording lengths. Prior work has shown that stochastic sparse sampling effectively handles variable-length signals, while retrieval-augmented approaches improve
explainability and robustness to noise and weak temporal correlations. In this study, we generalize the stochastic sparse sampling framework for retrieval-informed classification. Specifically, we weight window predictions by within-channel similarity and aggregate them in probability space, yielding convex series-level scores and an explicit evidence trail for explainability. Our method achieves competitive iEEG classification performance and provides practitioners with greater transparency and explainability. We evaluate our method in iEEG recordings collected in four medical centers, demonstrating its potential for reliable and explainable clinical variable-length time series classification.

\end{abstract}

\section{Introduction}

Artificial intelligence (AI) is increasingly embedded across clinical and translational workflows, with reported benefits for diagnostics, treatment planning, monitoring, and population health \cite{anwer_opportunities_2024, li_commercialization_2025, rajpurkar_ai_2022}. Nevertheless, routine deployment remains uneven. Two persistent barriers are the heterogeneity of clinical data and the need for transparent, clinician-oriented explanations \cite{sun_explainable_2025}.

One domain where the challenges are most keenly felt is medical-time series classification. Heart rate, glucose, and electrophysiology are examples of physiological signals that are both sparse and prone to noise, with their durations differing significantly between persons and events \cite{walther_systematic_2023, agliari_detecting_2020}. 
Most of the time series classification (TSC) research, however, remains centered on approaches that operate with fixed-length sequences only \cite{fawaz_deep_2019, foumani_deep_2023, mootoo_stochastic_2024}.

Recently, \citet{mootoo_stochastic_2024} proposed the Stochastic Sparse Sampling (SSS) to address variable-length time series classification (VTSC). SSS samples fixed-length windows from long recordings, computes local predictions using a backbone model, and aggregates these to obtain a series-level decision \cite{mootoo_stochastic_2024}. While effective and computationally tractable, SSS aggregates window predictions uniformly, thereby treating all sampled segments as equally informative; its explainability relies primarily on visualizations of local scores. 
This assumption might especially be problematic in real-world time series, however, where non-stationary, irregular patterns may occur infrequently and lack strong temporal correlations, making generalization difficult \cite{kim_reversible_2021, weigend_nonlinear_1995}. 

Retrieval-augmented methods address this by selectively leveraging similar past instances rather than memorizing all patterns. 
In time series, retrieval has also been explored across entities, where similarities guide aggregation for forecasting \cite{iwata_few-shot_2020, yang_mqretnn_2022}. Most recently, Retrieval-Augmented Forecasting of Time series (RAFT) \cite{han_retrieval_2025} introduces a similarity-based retrieval mechanism for forecasting: it retrieves past patches most similar to the current input and leverages their future continuations to improve predictions, with notable gains for rare patterns and weak temporal correlations

\cite{han_retrieval_2025}. However, RAFT is tailored to forecasting and does not directly address TSC.
Motivated by these advances, 
we propose RAxSS: Retrieval-Augmented Sparse Sampling for
Explainable Variable-Length Medical Time Series
Classification, a variable-length time series classification (VTSC) framework that integrates a retrieval-informed relevance computation into the SSS pipeline. Using a medical use case, we tackle the Seizure Onset Zone (SOZ) localization problem. More details about the problem can be found in the Appendix \ref{apdx:soz}. 

\method retains SSS’s stochastic, length-proportional sampling and replaces uniform averaging with a similarity-weighted convex mix of window predictions. Using pearson or cosine similarity (as in \citet{han_retrieval_2025}), each window’s top-$m$ within-series neighbors define a support score that is softmax-normalized into aggregation weights.
This design amplifies informative segments and downweights noisy ones. It also enables drill-down explanations: the series score is an additive sum of window contributions, each justified by a within-channel retrieval leaderboard.

\paragraph{Our contributions.}
 (i) We provide a \textbf{methodological advance} for variable-length time-series classification by introducing a retrieval-weighted aggregation mechanism that ranks and weights windows within each series, thereby improving uniform averaging while preserving the efficiency of stochastic sampling.
 (ii) We align \textbf{explanation with aggregation} through a weighting scheme that produces quantitative, window-level attributions. These extend beyond static heatmaps and enable principled drill-down from series-level predictions to segment-level contributions. 
 (iii) We demonstrate robustness in challenging regimes (the settings where retrieval excels for time-series modeling \cite{han_retrieval_2025}), while retaining compatibility with diverse backbones, including transformer variants.
 \hfill \break  
 \method adapts retrieval mechanisms 
 % originally 
 developed for forecasting the task of variable-length classification, combining stochastic coverage with similarity-guided prioritization to establish a framework that is selective, explainable, and well-suited to the demands of clinical time-series analysis.

\begin{figure}[t]
  \centering
  \begin{subfigure}{0.48\linewidth}
    \centering
    \includegraphics[width=\linewidth]{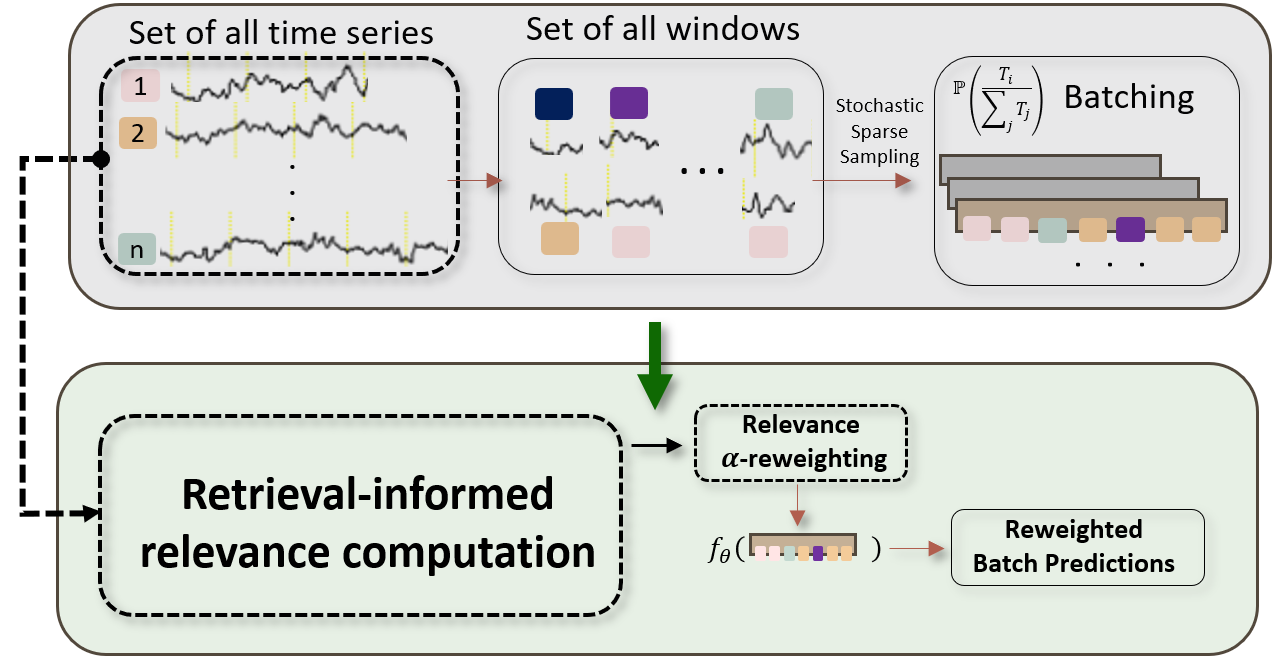}
    \caption{\textbf{RAxSS pipeline.} Green box: conceptual addition to SSS (gray). Dashed boxes: retrieval steps.}
    \label{fig:r2a-arch}
  \end{subfigure}\hfill
  \begin{subfigure}{0.48\linewidth}
    \centering
    \includegraphics[width=\linewidth]{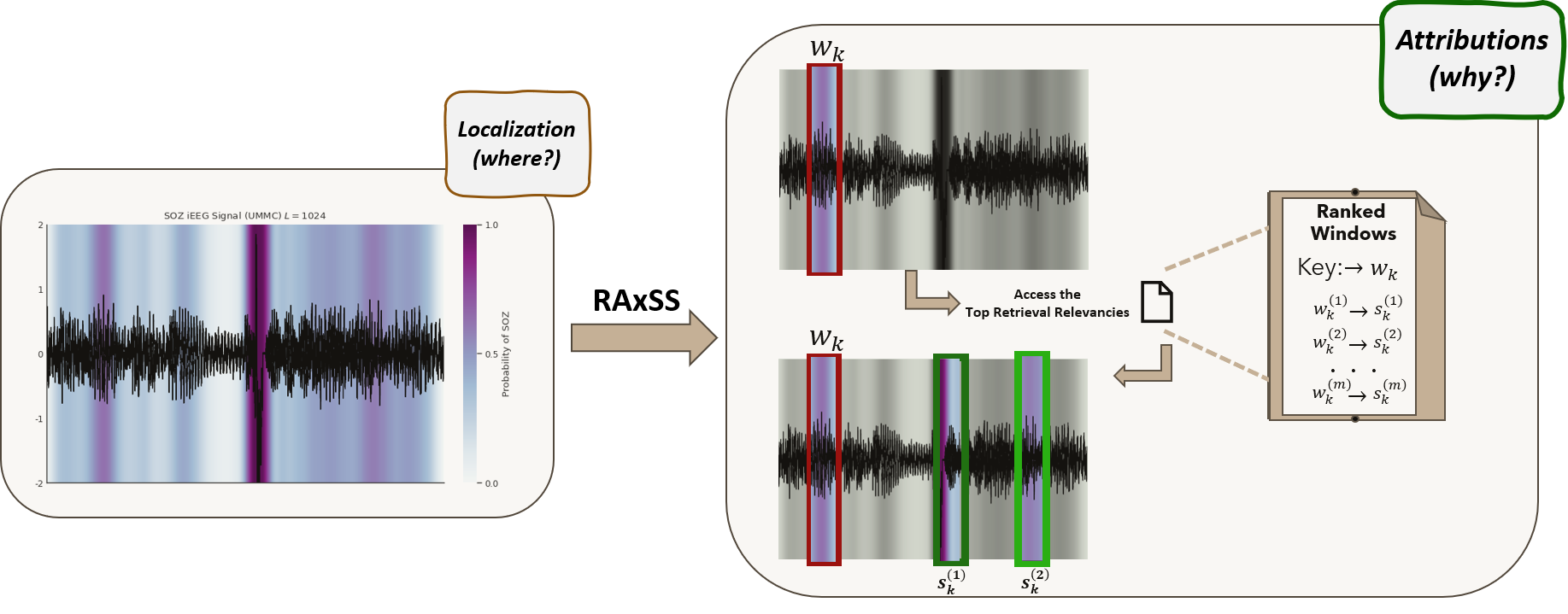}
    \caption{\textbf{Window ranking \& attribution.} Ranked, nonidentical neighbors explain each window’s influence.}
    \label{fig:r2a-xai}
  \end{subfigure}
  \caption{\method workflow: (a) end-to-end pipeline and (b) retrieval-weighted explainable module.}
  \label{fig:r2a}
\end{figure}

\section{Method}

\subsection{Datasets}

\paragraph{Epilepsy iEEG Multicenter Dataset}
We use the Epilepsy iEEG Multicenter Dataset, comprising of intracranial EEG (iEEG) recordings with seizure onset zone (SOZ) 
% annotations
from four centers: Johns Hopkins Hospital (JHH), the National Institutes of Health (NIH), University of Maryland Medical Center (UMMC), and University of Miami Jackson Memorial Hospital (UMH). Following the evaluation practice of \citep{mootoo_stochastic_2024}, we report F1 score, area under the curve (AUC), and accuracy as primary metrics. Summary statistics and additional dataset and data-preprocessing details are provided in Appendix \ref{apdx:dataset}.

\subsection{Framework overview}
\label{sec:framework}

RAxSS builds on SSS to handle variable-length medical time series by unifying sampling, retrieval, and aggregation in a single loop. As outlined in Fig.~\ref{fig:r2a-arch} and implemented in Alg.~\ref{alg:sss-retrieval}, long, noisy recordings are segmented into fixed-length windows, sampled \emph{length-proportionally} so that the probability of drawing from series $i$ is $p_i \propto T_i/\sum_j T_j$, and scored by a backbone $f_\theta$. In parallel (see Fig.~\ref{fig:r2a-xai}), a \emph{within-series} retrieval computes Pearson or cosine similarities, forms for each query window the top-$m$ \emph{nonidentical} neighbors (which may temporally overlap due to sliding extraction), and summarizes their support. The retrieval-aware aggregator (Alg.~\ref{alg:aggregate}) then converts these supports into softmax weights across windows and produces a \emph{convex} series-level prediction by re-weighting and aggregating window-level outputs. Fig.~\ref{fig:r2a-xai} further illustrates the explanatory consequence: each influential window is accompanied by a ranked, possibly overlapping but nonidentical set of neighbors whose weights quantify \emph{why} it matters. The result maintains SSS efficiency while incorporating resilient, retrieval-guided weighting and a clear evidence path from windows to the final prediction.

\subsection{Enhancing explainability with \method}

A consequence of applying RAxSS is explainability, where we go beyond just localization but can also access the attributions. In more detail, for series $i$ with window index set $K_i$, the base model outputs window
posteriors $p_k=\mathrm{softmax}(z_k)\in\Delta^{C-1}$, $k\in K_i$.

For each $k\in K_i$, retrieve the $m$ most similar nonidentical windows from
the same series under $\phi\in\{\text{Pearson},\text{Cosine}\}$:
\begin{equation}
\label{eq:neighbor-sim}
\begin{aligned}
s_k^{(j)} &= \phi\!\big(w_k,w_j\big), && j\in N_k,\quad |N_k|=m,
\bar s_k &= \frac{1}{m}\sum_{j\in N_k} s_k^{(j)} \, 
\end{aligned}
\end{equation}

Subsequently, we define window influence weights via a temperatured softmax over $\{\bar s_t\}_{t\in K_i}$:
\begin{equation}
\label{eq:alpha-softmax}
\alpha_k \;=\; \frac{\exp(\bar s_k/\tau)}{\sum_{t\in K_i}\exp(\bar s_t/\tau)} \in [0,1],
\qquad \sum_{k\in K_i}\alpha_k=1 \, .
\end{equation}

\paragraph{Aggregation in probability space.}
Series-level probabilities are a convex combination of window posteriors (proof in Appendix \ref{apdx:convex}):
\begin{equation}
\label{eq:series-convex-comb}
\hat p^{(i)}=\sum_{k\in K_i}\alpha_k\,p_k \, .
\end{equation}

\paragraph{From "where?" to "why?"}
Explainability should go beyond localization and provide reasons for \emph{why} specific regions are trusted. In Fig.~\ref{fig:r2a-xai}, the left panel shows the window-probability heatmaps of \citet{mootoo_stochastic_2024}, which indicate \emph{where} the model is confident. RAxSS adds attributions to answer the \emph{why}: for each influential window $k$, we expose the evidence used to compute its weight.
% qs: weight is lambda k or aplha k ? it is the same right. aj: Yes, I changed it to be alpha so it is notation compatible with the literature's work. Thank you.
$\alpha_k$ by reporting (i) its summary support $\bar{s}_k$, the mean similarity to its top-$m$ within-series neighbors, and (ii) a ranked neighbor leaderboard $\{(w_k^{(j)},\,s_k^{(j)}): j\in N_k\}$ with timestamps. These quantities explain why window $k$ received a high contribution 
% qs: here still \lambda_k. pls change aj: changed. 
$\alpha_k\,p_{k,c}$ to the final series-level probability.
%Trivially, s
Since
\begin{equation}
\label{eq:alpha-derivative}
\frac{\partial \alpha_k}{\partial s_k^{(j)}} \;=\; \frac{1}{m\tau}\,\alpha_k\big(1-\alpha_k\big) \;>\; 0,
\end{equation}
increasing any neighbor similarity strictly increases $\alpha_k$ (holding all
other $\bar s_t$ fixed), making the leaderboard a faithful explanation of why
$w_k$ was weighted highly. For a selected channel we overlay: (a) the raw signal, (b) the window probability heatmap (\emph{localization}), and (c) for the top-$m$ supporting windows and their support $\alpha_{k,j}$ values (\emph{attribution}). See Fig.~\ref{fig:r2a-xai} for an example visualization of our proposed explainability framework.

\section{Experiments \& Results}
\begin{table}[t]
\centering
\caption{SOZ localization on \textbf{All} centers. F1, AUC, and Accuracy are averaged over \textbf{5 seeds}. 
For our runs (\emph{RAxSS} variants and \emph{SSS (reproduction)}), we used the \emph{same seed set} and backbone code; the line \emph{SSS (paper)} is the value reported by the original authors. 
Boldface values with $^{\ast}$ and $^{\dagger}$ denote the best and second-best results per column, respectively.}

\label{tab:experiment-results}
\begin{tabular}{lccc}
\toprule
\textbf{Model} & \textbf{F1} & \textbf{AUC} & \textbf{Acc.(\%)} \\
\midrule
\textbf{RAxSS (cosine)} & 0.6967 $\pm$ 0.0791 & \textbf{0.8046}$^{\ast}$ $\pm$ 0.0346 & 69.76 $\pm$ 5.25 \\
\textbf{RAxSS (pearson)} & \textbf{0.7275}$^{\dagger}$ $\pm$ 0.0489 & 0.7980 $\pm$ 0.0537 & \textbf{70.51}$^{\dagger}$ $\pm$ 3.59 \\
\textbf{SSS (reproduction)} & \textbf{0.7437}$^{\ast}$ $\pm$ 0.0537 & \textbf{0.8035}$^{\dagger}$ $\pm$ 0.0686 & \textbf{71.14}$^{\ast}$ $\pm$ 6.31 \\
\midrule
\textbf{SSS (\citet{mootoo_stochastic_2024})} & 0.7629 & 0.7999 & 72.35 \\
PatchTST (\citet{nie_time_2023}) & 0.7097 & 0.7852 & 66.83 \\
TimesNet (\citet{wu_timesnet_2023}) & 0.6897 & 0.7174 & 65.98 \\
ModernTCN (\citet{luo_moderntcn_2024}) & 0.6938 & 0.7305 & 68.42 \\
DLinear (\citet{zeng_are_2023}) & 0.6916 & 0.7044 & 68.41 \\
ROCKET (\citet{dempster_rocket_2019}) & 0.6847 & 0.7481 & 69.27 \\
Mamba (\citet{gu_mamba_2024}) & 0.6452 & 0.7134 & 64.39 \\
GRUs (\citet{bahdanau_neural_2016}) & 0.6948 & 0.7340 & 65.85 \\
LSTM (\cite{hochreiter_long_1997}) & 0.6709 & 0.7144 & 65.43 \\
\bottomrule

\end{tabular}
\end{table}

\paragraph{Results}
On multicenter iEEG, RAxSS is competitive with strong baselines (Table \ref{tab:experiment-results}). The cosine variant achieves the best AUC (0.8046 $\pm$ 0.0346), edging the reproduced SSS (\(0.8035 \pm 0.0686\)) and outperforming non-SSS baselines (e.g., PatchTST \(0.7852\)). The Pearson variant yields higher F1 than cosine (0.7275 $\pm$ 0.0489 vs.\ 0.6967 $\pm$ 0.0791) and strong accuracy (\(70.51 \pm 3.59\)), close to SSS (\(71.14 \pm 6.31\)). Overall, cosine favors AUC, while Pearson offers a better F1/accuracy trade-off, letting practitioners pick the similarity to prioritize discrimination or balanced detection, while retaining built-in explainability. The training details are provided in Appendix \ref{tab:hyperparams}.

\section{Discussion \& Conclusion}

In this paper, our primary goal was to provide a more clinician-oriented, steerable and explainable framework for VTSC. To achieve this, we: (i) coupled stochastic sparse sampling with within-recording retrieval and probability-space aggregation; (ii) made explanations by exposing additive window contributions and an evidence leaderboard for influential windows; and (iii) preserved practicality via a model-agnostic, privacy-friendly design with simple knobs for steering. Results showed robust, competitive performance across centers, all while maintaining more transparency and explainability. RAxSS consistently ranks among the top approaches across metrics and sites, and we expect routine calibration and hyperparameter tuning to further boost absolute performance.

The window-based design already gives granular localizations by overlaying window-level probabilities (the \emph{where}). 
To explain the \emph{why}, we present, for each influential window, a ranked list of its (top-$m$) within-recording neighbors (nonidentical, overlap allowed) with their similarity scores and resulting weights. This is justified because a) similarities determine the window’s weight via mean support and b) the cross-window softmax is strictly increasing in that support. Thus the same evidence that raises a window’s weight justifies its contribution, yielding a faithfulness-oriented "why", on top of "where". Despite this transparency, finer-grained, mechanistic explanations will require probing internal representations and decision pathways.

In clinical use, inference is typically per recording, so length-proportional sampling offers no test-time benefit. Retrieval remains pivotal: it reweights window predictions by agreement with within-recording neighbors, improving robustness to noisy and idiosyncratic windows and providing inspectable evidence via the neighbor leaderboard. Over this medical setting, we couple the retrieval concept into time series classification (prior work emphasized forecasting \citep{han_retrieval_2025}) enabling domain-aligned control to match clinical priorities.

\paragraph{Future work.}
Our current implementation performs retrieval and aggregation strictly within the same channel/recording. This choice (i) avoids dependence on cross-subject/center labels, (ii) reduces privacy exposure by not querying external data, and (iii) keeps the approach generic for other clinical time-series tasks. A natural extension is \emph{pattern-level} retrieval: indexing canonical events (e.g., seizure onsets) and retrieving neighbors from the same subject or a curated, cross-center repository. While this may strengthen the quality of evidence and enable case-based reasoning, it requires additional curation/metadata and stronger governance (privacy and access control). Beyond scope, two technical directions are promising: learning the similarity/temperature parameters from data, and conducting comprehensive faithfulness stress tests (e.g., deletion/insertion tests, retrieval randomization, and counterfactual probes) to further validate the explanations.

%%%%%%%%%%%%%%%%%%%%%%%%%%%%%%%%%%%%%%%%%%%%%%%%%%%%%%%%%%%%
\newpage
\bibliographystyle{unsrtnat} % numeric, unsorted -> sorted by appearance
\bibliography{references, references-2}
\newpage
\appendix

\section{Technical Appendices and Supplementary Material}

\subsection{Proposition (convexity of the series-level probabilities).}
\label{apdx:convex}

Let $K_i$ be the set of windows for series $i$. Assume each window posterior
$p_k\in\Delta^{C-1}$ (entries nonnegative and summing to $1$). Define
\[
\alpha_k=\frac{\exp(\bar s_k/\tau)}{\sum_{t\in K_i}\exp(\bar s_t/\tau)},\qquad \tau>0.
\]
Then $\hat p^{(i)}=\sum_{k\in K_i}\alpha_k\,p_k\in\Delta^{C-1}$, i.e., it is a
\emph{convex combination} of $\{p_k\}$.

\emph{Proof.}
Since $\exp(\cdot)>0$, we have $\alpha_k\ge 0$ for all $k$, and by construction
$\sum_{k\in K_i}\alpha_k=1$. For each class $c$,
\[
\hat p^{(i)}_c=\sum_{k}\alpha_k\,p_{k,c}\ \ge\ 0
\quad\text{because}\quad \alpha_k,p_{k,c}\ge 0.
\]
Moreover,
\[
\sum_{c=1}^C \hat p^{(i)}_c
= \sum_{c}\sum_{k}\alpha_k\,p_{k,c}
= \sum_{k}\alpha_k \big(\sum_{c}p_{k,c}\big)
= \sum_{k}\alpha_k \cdot 1
= 1 .
\]
Thus $\hat p^{(i)}$ has nonnegative entries summing to $1$, so
$\hat p^{(i)}\in\Delta^{C-1}$ and, by definition, is a convex combination of the
$\{p_k\}$.  \hfill$\square$

\subsection{Seizure Onset Zone (SOZ) Localization problem description}
\label{apdx:soz}

Developing explainable methods for variable-length time series classification (VTSC) is especially critical in seizure onset zone (SOZ) localization, where clinicians must determine the brain regions that initiate seizures \cite{balaji_seizure_2022}. Epilepsy affects over 50 million people worldwide, making it one of the most prevalent but still poorly characterized neurological conditions \cite{noauthor_epilepsy_nodate, stafstrom_seizures_2015, mootoo_stochastic_2024}. For nearly one-third of patients, medication is ineffective, leaving surgery as the only option and placing high demands on accurate SOZ mapping. Current practice involves surgically implanting electrodes in candidate regions and visually inspecting intracranial EEG (iEEG) recordings to classify which channels correspond to the SOZ.

\newpage
\subsection{Algorithms}
\label{apdx:algo}

\begin{algorithm}[ht]
\caption{Variable Length Time Series Training Algorithm with Retrieval-augmented Aggregation (Single Epoch)}
\label{alg:sss-retrieval}
\DontPrintSemicolon
\SetKwInOut{Input}{Input}
\SetKwInOut{Output}{Output}

\Input{%
Time series $\mathcal{X}=\{(x^{(1)}_t)_{t=1}^{T_1},\dots,(x^{(n)}_t)_{t=1}^{T_n}\}$; \\
Labels $\mathcal{Y}=\{y^{(1)},\dots,y^{(n)}\}$; model $f_\theta$; batch size $B$; loss $\mathcal{L}$; .
}
\Output{Updated parameters $\theta$}

$\mathcal{W} \leftarrow$ set of \emph{all} windows from each series in $\mathcal{X}$\;

\While{$\mathcal{W}\neq \emptyset$}{
  \tcp{Sample a minibatch of windows with length-proportional probabilities}
  $\mathcal{W}_0 \leftarrow \textsc{Sample}(\mathcal{W}, B)$ with
  $\Pr(\text{series } i) = \dfrac{T_i}{\sum_j T_j}$\;

  \For{$i = 1,\dots,n$}{
    $\mathcal{W}_i \leftarrow \{\, w \in \mathcal{W}_0 \mid w \text{ comes from series } i \,\}$\;
    
    % $\mathcal{A} \leftarrow [\;]$ list to be stored for retrieval based aggregation; 
    
    \If{$\mathcal{W}_i = \emptyset$}{\textbf{continue}}
    
    \tcp{Per-window retrieval signals for series $i$}
    \ForEach{$w_k \in \mathcal{W}_i$}{
      $R_i[k] \leftarrow \textsc{Retrieve}(w_k, T_i)$\;
      \tcp*[f]{a dictionary $\{k^{(1)}_{i}\!:\!\rho^{(1)}_{i}, \ldots, k^{(m)}_{i}\!:\!\rho^{(m)}_{i}\}$ of Pearson | Cosine scores}
    }
    
    \tcp{Window-level predictions}
    $\mathcal{Y}_i \leftarrow \{\, f_\theta(w) \mid w \in \mathcal{W}_i \,\}$\;
    
    \tcp{Aggregate window predictions using retrieval signals}
    $\hat{y}^{(i)} \leftarrow \textsc{Aggregate}(\mathcal{Y}_i, R_i)$\;
  }

  \tcp{Batch loss over (non-empty) series present in $\mathcal{W}_0$}
  %qs: This will include windows that are not included in the average denominator if you use 1/n. 
  % $\mathcal{L}_{\text{batch}} \leftarrow
  %   \dfrac{1}{n} \sum_{i=1}^{n}   \mathcal{L}\!\left(\hat{y}^{(i)},\, y^{(i)}\right)$\;
    % qs: added

  $I \leftarrow \{\, i \in \{1,\dots,n\} \mid \mathcal{W}_i \neq \varnothing \,\}$
  
  $\mathcal{L}_{\text{batch}} \leftarrow
  \dfrac{1}{|I|}{\displaystyle\sum\limits_{i \in I}}\mathcal{L}\!\left(\hat{y}^{(i)},\, y^{(i)}\right)$
  
  \tcp{Parameter update}
  $\theta \leftarrow \textsc{Update}(\theta,\, \mathcal{L}_{\text{batch}})$\;

  \tcp{Remove sampled windows from the pool}
  $\mathcal{W} \leftarrow \mathcal{W} \setminus \mathcal{W}_0$\;
}
\Return{$\theta$}
\end{algorithm}

\begin{algorithm}[H]
\caption{\textsc{Aggregate}}
\label{alg:aggregate}
\DontPrintSemicolon
\SetKwInOut{Input}{Input}
\SetKwInOut{Output}{Output}

\Input{%
  Windows for series $i$: $\mathcal{W}_i$; \\
   $\mathcal{Y}_i \leftarrow \{\, f_\theta(w) \mid w \in \mathcal{W}_i \,\}$\ \quad \tcp{Window-level predictions}
 \\
  Retrieval map $R_i$ with $R_i[k]=\big(s_k^{(1)},\ldots,s_k^{(m)}\big)$ \quad 
  \tcp{top-$m$}\\
  Temperature $\tau>0$
}
\Output{Series-level probability $\hat{y}^{(i)} \in \mathbb{R}^C$}

\BlankLine

\tcp{1) Summarize neighbor support per window}
\ForEach{$k \in \mathcal{W}_i$}{
  $\bar{s}_k \leftarrow \dfrac{1}{m}\sum_{j=1}^{m} s_k^{(j)}$ \quad \quad
  \tcp{mean similarity for window $k$}
}

\BlankLine
\tcp{2) Softmax weights across windows}
\ForEach{$k \in \mathcal{W}_i$}{
  $a_k \leftarrow \exp\!\big(\bar{s}_k/\tau\big)$
}
$Z \leftarrow \sum_{t \in \mathcal{W}_i} a_t$\;

\ForEach{$k \in \mathcal{W}_i$}{
  $\alpha_k \leftarrow a_k / Z$ \quad \quad \quad \tcp{$\alpha_k \ge 0$, $\sum_k \alpha_k = 1$}
}

\BlankLine

\tcp{3) Aggregate in probability space}
$\hat{p}^{(i)} \leftarrow \sum_{k \in \mathcal{W}_i} \alpha_k \, \mathcal{Y}_{ik}$\;

\BlankLine
\Return{$\hat{y}^{(i)}$}
\end{algorithm}

\section{Dataset and Preprocessing}
\subsection{Dataset}\label{apdx:dataset}
Following the protocol of \cite{mootoo_stochastic_2024}, we use a multicenter iEEG cohort with clinical annotations of the seizure onset zone (SOZ). For each site, we report the number of patients recorded ($n$), the number with SOZ labels ($n_{\text{SOZ}}$), the total number of channel time series ($n_{\text{ts}}$), the proportion of SOZ labeled ($p_{\text{SOZ}}$), the iEEG modality, nominal sampling frequency, and availability of postoperative outcome labels. A summary is provided in Table~\ref{tab:dataset-summary}.
\clearpage
\begin{table}[t]
\centering
\caption{Multicenter iEEG summary. $n$: patients recorded; $n_{\text{SOZ}}$: patients with SOZ annotation; $n_{\text{ts}}$: channel time series; $p_{\text{SOZ}}$: fraction of series labeled SOZ.}
\label{tab:dataset-summary}
\begin{tabular}{lrrrrlll}
\toprule
\textbf{Medical Center} & \textbf{$n$} & \textbf{$n_{\text{SOZ}}$} & \textbf{$n_{\text{ts}}$} & \textbf{$p_{\text{SOZ}}$} & \textbf{iEEG Type} & \textbf{Freq (Hz)} & \textbf{Outcomes} \\
\midrule
JHH  & 7  & 3  & 1458 & 7.48\%  & ECoG & 1000      & No  \\
NIH  & 14 & 11 & 3057 & 12.23\% & ECoG & 1000      & Yes \\
UMMC & 9  & 9  & 2967 & 5.56\%  & ECoG & 250--1000 & Yes \\
% qs: UMH? typo maybe, aj: yes.
UMH  & 5  & 1  & 129  & 25.58\% & ECoG & 1000      & No  \\
\bottomrule
\end{tabular}
\end{table}

\noindent
Per \citet{mootoo_stochastic_2024}, we filter to patients with SOZ annotations when forming the supervised subsets ($n_{\text{SOZ}}$). Because SOZ vs.\ non-SOZ is highly imbalanced at the series level, later class balancing reduces the effective number of training/validation examples for each site.

\subsection{Data preprocessing}\label{apdx:data}
Unless stated otherwise, we largely adhere to \cite{mootoo_stochastic_2024}. Each patient contributes multiple channels (electrodes). For every site we:
\begin{enumerate}
    \item Extract all channels and form per-channel univariate time series;
    \item Perform class balancing so that SOZ and non-SOZ series counts are equal within the training/validation splits (non-SOZ downsampling);
    \item Split channels into train/validation/test at approximately 70\% /10\% /20\%, ensuring no temporal leakage across splits during window sampling;
    \item Z-score normalize each channel independently to zero mean and unit variance.
\end{enumerate}
We report F1, AUC, and accuracy in the main results.

\subsection{Reproducibility \& Hyperparameters}
\label{sec:repro}

All training hyperparameters are listed in Table~\ref{tab:hyperparams}. Each experiment is run with five fixed seeds (69421–69425). We will release the full codebase, configuration files, and run scripts in a public repository at camera-ready, including exact commands and environment specifications to reproduce Table~\ref{tab:experiment-results}.

\begin{table}[h]
\centering
\small
\caption{RAxSS hyperparameters and data settings.}
\label{tab:hyperparams}
\begin{tabular}{ll}
\toprule
\multicolumn{2}{c}{\textbf{Experiment / Reporting}} \\
\midrule
Model ID & \texttt{PatchTSTBlind} \\
Seeds & \([69421, 69422, 69423, 69424, 69425]\) \\
Learning type & \texttt{sl} (supervised) \\
Metrics \& selection & report \texttt{acc}, \texttt{ch\_acc}, others; tune on \texttt{ch\_f1}; select on \texttt{ch\_acc} \\
Task & \texttt{classification} \\
GPU & \texttt{gpu\_id=0}; single‐GPU runs (see \ref{apdx:compute}) \\
\midrule
\multicolumn{2}{c}{\textbf{Data / Sampling / Preprocess}} \\
\midrule
Dataset & \texttt{open\_neuro} (multicenter iEEG) \\
Split & train/val/test = 0.7/0.1/0.2, \emph{no temporal leakage}; class balancing in train/val \\
Windowing & length \(L{=}1024\), stride \(=5\); univariate channels (\(C{=}1\)) \\
Batching & length-proportional stochastic sparse sampling (SSS) \\
Resizing & \texttt{pad\_trunc}; \texttt{seq\_load=True}; \texttt{num\_workers=8} \\
Scaling & per‐channel z‐score; \texttt{scale=True}; \texttt{shuffle\_test=True} \\
\midrule
\multicolumn{2}{c}{\textbf{Backbone / Architecture}} \\
\midrule
Encoder layers & \texttt{num\_enc\_layers=2} \\
Dims / heads & \texttt{d\_model=32}, \texttt{d\_ff=128}, \texttt{num\_heads=4} \\
Dropout & \texttt{attn\_dropout=0.3}, \texttt{ff\_dropout=0.3}, \texttt{pred\_dropout=0.0} \\
Head & \texttt{linear} \\
RevIN & \texttt{revin=True}, \texttt{revin\_affine=True}, \texttt{revout=False} \\
\midrule
\multicolumn{2}{c}{\textbf{Retrieval \& Aggregation (RAxSS)}} \\
\midrule
Similarity & Pearson or cosine (within series/channel) \\
Support $\to$ weights & average top-\(m\) [10] similarities; softmax with temperature \(\tau>0\) across windows \\
Aggregation & In probability space (Alg.~\ref{alg:aggregate}) \\
Use relevance & \texttt{use\_relevance=True} \\
\midrule
\multicolumn{2}{c}{\textbf{Optimization / Training}} \\
\midrule
Epochs \& batch & \texttt{epochs=50}, \texttt{batch\_size=8192} \\
Optimizer & \texttt{adam}, \texttt{weight\_decay=1e-6} \\
Scheduler & cosine warmup: \texttt{warmup\_steps=100}, \texttt{T\_max=700}, \texttt{start\_lr=0.0}, \texttt{final\_lr=1e-6}, \texttt{max\_lr=3e-4} \\
Early stopping & \texttt{patience=5} \\
Loss & \texttt{BCE} (\texttt{ch\_loss=True}, type \texttt{BCE}, \(\alpha{=}0.0,\beta{=}1.0\)) \\
\midrule
\multicolumn{2}{c}{\textbf{Dataset‐specific (OpenNeuro settings)}} \\
\midrule
Kernels/pooling & \texttt{kernel\_size=24}, \texttt{kernel\_stride=-1}, \texttt{pool\_type=avg} \\
Centers & \texttt{all\_clusters=True} \\
Task & binary classification (\texttt{pred\_len=1}) \\
\bottomrule
\end{tabular}
\end{table}

\subsection{Computational Resources}
\label{apdx:compute}
Experiments were conducted on a single NVIDIA T4 GPU with 32\,GB system RAM, each training run (per seed) took about 1\,hour. All computations used PyTorch with CUDA \citep{paszke_pytorch_2019}.

\end{document}